\documentclass[letterpaper, 10 pt, conference]{ieeeconf}  

\IEEEoverridecommandlockouts                              

\usepackage{graphics} 
\usepackage{epsfig} 
\usepackage{mathptmx} 
\usepackage{times} 
\usepackage{amsmath} 
\usepackage{amssymb}  

\usepackage{graphics} 
\usepackage{epsfig} 
\usepackage{mathptmx} 
\usepackage{times} 
\usepackage{amsmath} 
\usepackage{amssymb}  
\usepackage{epsfig,color,subfigure,setspace}
\usepackage{amsfonts}
\usepackage{mathrsfs}
\usepackage{mathtools, amsmath, amssymb, commath}

\usepackage{graphicx}  
\usepackage{url}
\usepackage{array, tabularx}
\usepackage{multicol}  
\usepackage[dvipsnames]{xcolor} 
\usepackage{ragged2e} 
\usepackage{float}
\usepackage{enumerate}
\usepackage{cite, soul}
\usepackage{epsfig, color, subfigure, setspace}
\usepackage{epstopdf}
\usepackage{verbatim}
\usepackage{tikz}
\usepackage{fancyhdr}
\usepackage{setspace}
\usepackage{bigints}
\usepackage{mathtools, amsmath, amssymb, commath}
\usepackage{multirow}
\usepackage{eqnarray}
\usepackage{romannum}
\usepackage{algorithm}
\usepackage{gensymb}
\usepackage{hyperref}
\usepackage[noend]{algpseudocode}
\usepackage[noend]{algpseudocode}

\title{\LARGE \bf
A Data-Driven Model with Hysteresis Compensation for I$^2$RIS Robot  
}

\author{Mojtaba Esfandiari$^{1}$, Yanlin Zhou$^{1}$, Shervin Dehghani$^{2}$, Muhammad Hadi$^{1}$, Adnan Munawar$^{1}$, Henry Phalen$^{1}$,\\{\it Graduate Student Member, IEEE}, Peter Gehlbach$^{3}$, {\it Member, IEEE}, \\ Russell H. Taylor$^{4}$, {\it Life Fellow, IEEE} and Iulian Iordachita$^{1}$, {\it Senior Member, IEEE}
\thanks{*This work was supported by U.S. National Institutes of Health under the grants number 2R01EB023943-04A1 and 1R01 EB025883-01A1, and partially by JHU internal funds.}
\thanks{$^{1}$ Mojtaba Esfandiari, Yanlin Zhou, Adnan Munawar, Muhammad Hadi, Henry Phalen, and Iulian Iordachita are with the Department of Mechanical Engineering and also Laboratory for Computational Sensing and Robotics, Johns Hopkins University,
Baltimore, MD, 21218, USA. 
        ({\tt\small mesfand2,yzhou144,amunawa2,mhadi2,henry.phalen
        ,iordachita@jhu.edu})}%
\thanks{$^{2}$ Shervin Dehghani is with the Department of Computer Science, Technische Universit\"{a}t M\"{u}nchen, M\"{u}nchen 85748 Germany.
        ({\tt\small shervin.dehghani@tum.de})}
        \thanks{$^{3}$ Peter Gehlbach is with the Wilmer Eye Institute, Johns Hopkins Hospital, Baltimore, MD, 21287, USA. ({\tt\small pgelbach@jhmi.edu})
}%
\thanks{$^{4}$Russell H. Taylor is with the Department of Computer Science and also the Laboratory for Computational Sensing and Robotics at the Johns Hopkins University, Baltimore, MD, 21218, USA. ({\tt\small rht@jhu.edu})
}
}

\begin{document}

\maketitle
\thispagestyle{empty}
\pagestyle{empty}

\begin{abstract}

Retinal microsurgery is a high-precision surgery performed on an exceedingly delicate tissue. It now requires extensively trained and highly skilled surgeons. Given the restricted range of instrument motion in the confined intraocular space, and also potentially restricting instrument contact with the sclera, snake-like robots may prove to be a promising technology to provide surgeons with greater flexibility, dexterity, space access, and positioning accuracy during retinal procedures requiring high precision and advantageous tooltip approach angles, such as retinal vein cannulation and epiretinal membrane peeling. Kinematics modeling of these robots is an essential step toward accurate position control, however, as opposed to conventional manipulators, modeling of these robots does not follow a straightforward method due to their complex mechanical structure and actuation mechanisms. Especially, in wire-driven snake-like robots, the hysteresis problem due to the wire tension condition can have a significant impact on the positioning accuracy of these robots. In this paper, we proposed an experimental kinematics model with a hysteresis compensation algorithm using the probabilistic Gaussian mixture models (GMM) Gaussian mixture regression (GMR) approach. Experimental results on the two-degree-of-freedom (DOF) integrated robotic intraocular snake (I$^2$RIS) show that the proposed model provides $0.4\degree$ accuracy, which is overall 60$\%$ and 70$\%$ of improvement for yaw and pitch degrees of freedom, respectively, compared to a previous model of this robot.                


\end{abstract}

\section{INTRODUCTION} 

Retinal vein occlusion (RVO) is one of the most prevalent retinovascular diseases. It occurs because of the existence of an obstacle in the retinal vein blood flow. The obstacle may be a clot due to systemic diseases like cardiovascular disease, hypertension, diabetes mellitus, hyperlipidemia, or stroke \cite{rogers2010prevalence, scott2022retinal}. Retinal vein cannulation (RVC) is a potential surgical intervention for RVO which requires advanced surgical skills due to the delicate and fragile nature, as well as the small size, of retinal veins ($\phi<100$ $\mu m$) \cite{vander2020robotic}. A major challenge during RVC is to accurately maintain a cannula inside the vein to perform prolonged perfusion with a therapeutic agent. This is in part due to physiological hand tremor (with root mean square RMS amplitude in the range of 50-200 $\mu m $), and patient movement \cite{wells2013comparison}.
\begin{figure}[t!]
    \centering
    \includegraphics[width=0.49\textwidth]{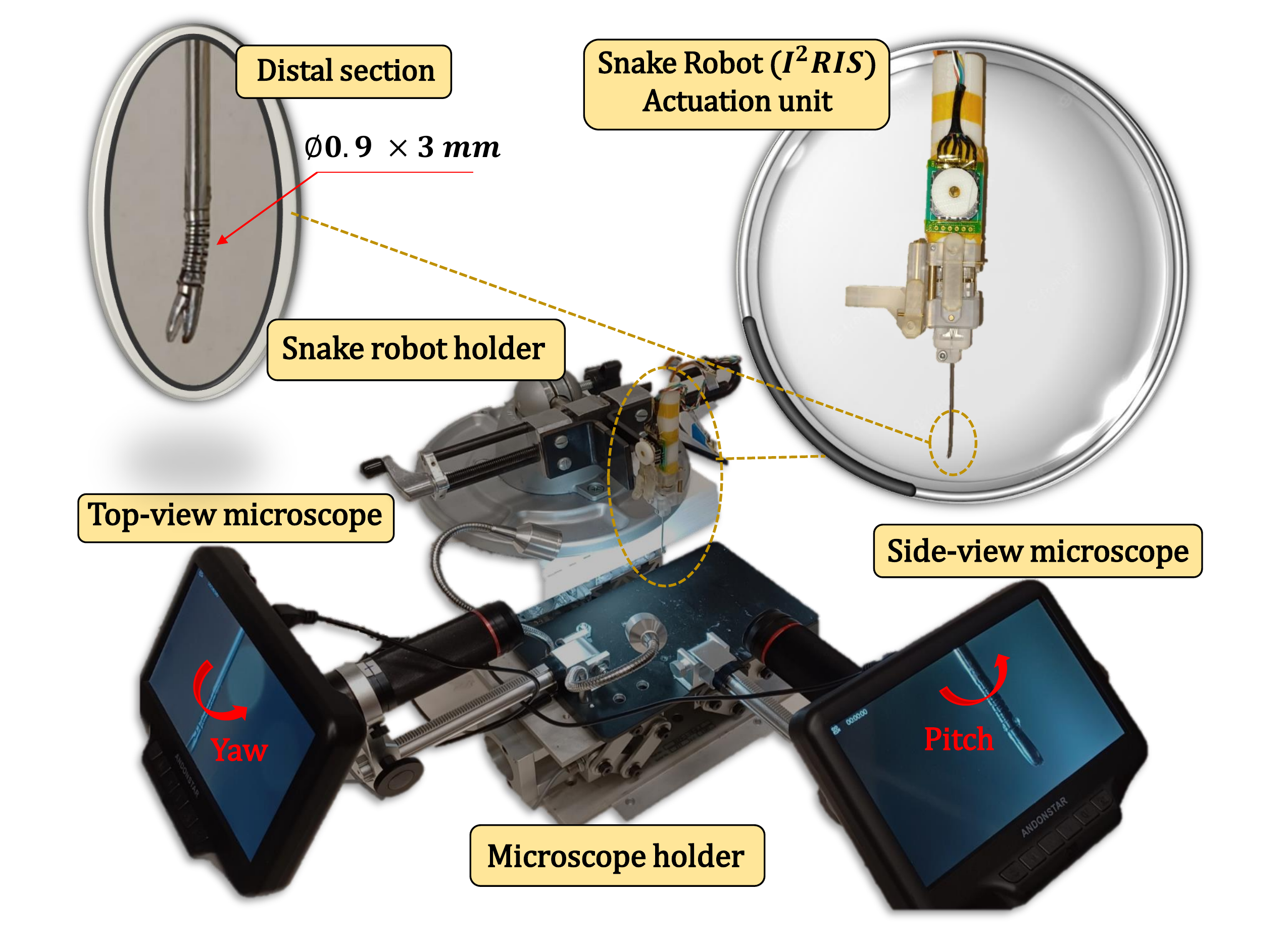}
      \caption{ Experiment setup: the I$^2$RIS snake robot and its actuation unit, the top-view and side-view microscopes to measure yaw and pitch angles.}
      \vspace{-0.5cm}
      \label{fig: experimet_setup}
   \end{figure}
   
To remove hand tremors and improve positioning accuracy, several robotic systems have been developed such as the Steady Hand Eye Robot (SHER) \cite{uneri2010new}, Intraocular Robotic Interventional Surgical System (IRISS) \cite{wilson2018intraocular}, PRECEYES Surgical System (Preceyes B.V., NL) \cite{PRECEYES2021robotic}, and others \cite{gijbels2013design, tanaka2015quantitative, nasseri2013introduction}. The SHER is a 5-DOF admittance-type robot manipulator capable of providing surgical instrument manipulation in a cooperative control mode between the surgeon and robot end-effector.

\begin{figure}[t!]
    \centering
    \includegraphics[width=0.46\textwidth]{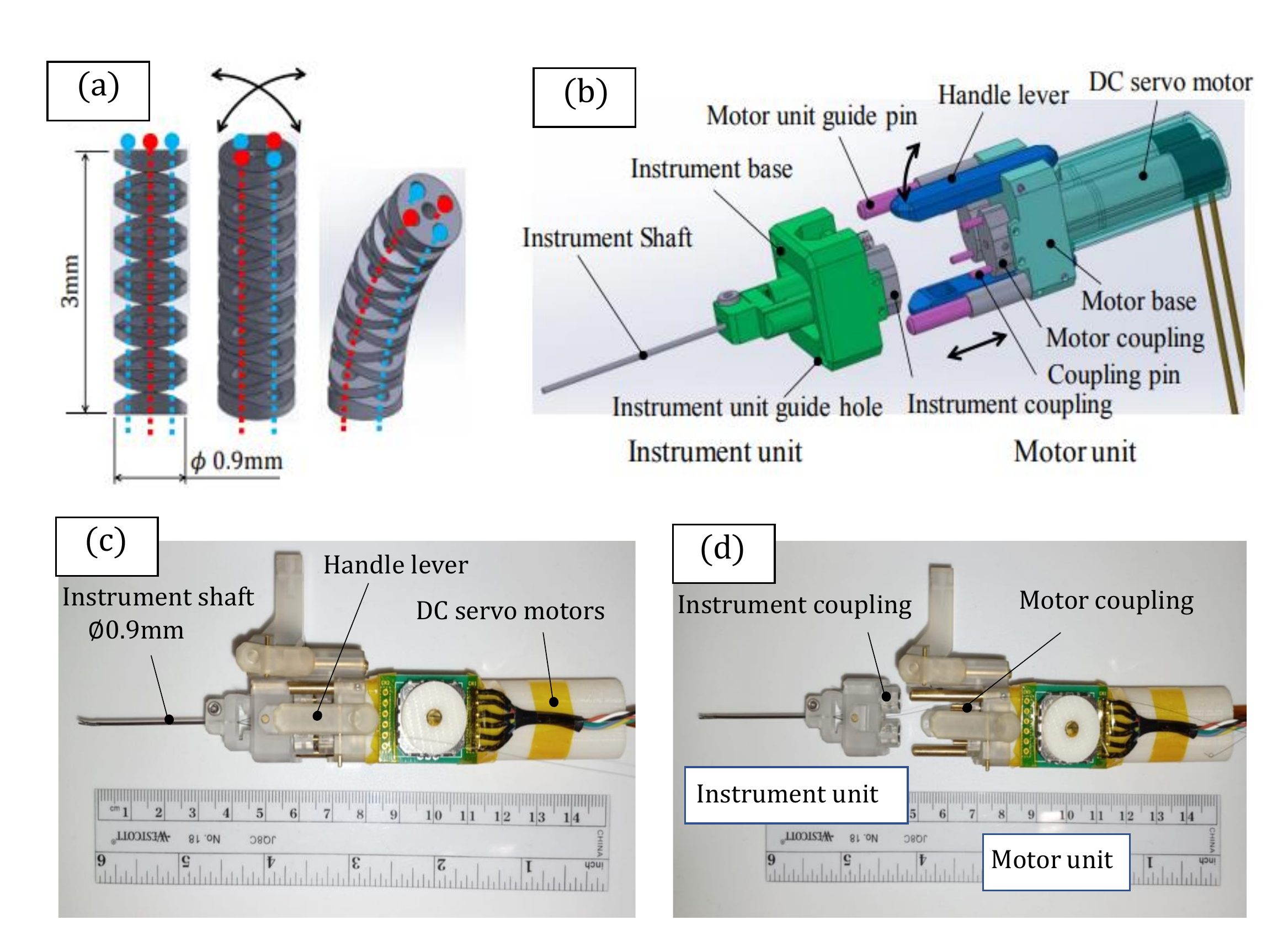}
      \caption{  Overview of the I$^2$RIS robot. (a) Schematic design of dexterous distal section, (b) detachable instrument unit from the actuation unit, (c) actual robot in the attached mode, (d) actual robot in the detached mode \cite{jinno2020improved}.}
       \vspace{-0.5cm}
      \label{fig: I2RIS_model}
   \end{figure}

Adding extra DOF at the distal end of an instrument can improve a robot's capability for procedures such as epiretinal membrane (ERM) peeling. These procedures require a higher level of dexterity and can thus benefit from optimized approach angles. Therefore, adding a wrist-like mechanism such as the Integrated Robotic Intraocular Snake (IRIS), at the distal end of the SHER is potentially useful \cite{he2015iris}. The second generation of this robot, i.e. the I$^2$RIS robot (Fig. \ref{fig: experimet_setup}), benefits from a smaller actuation unit with a detachable instrument section that makes it more compatible with the SHER robot end-effector and is also suitable for sterilization in clinical applications (Fig. \ref{fig: I2RIS_model})\cite{jinno2020improved}. 

To perform autonomous position control or teleoperation with virtual fixtures using a snake robot, it is necessary to have a good model for input-output mapping that will enhance operation accuracy and performance. Therefore, several analytical and experimental methods for kinematics modeling in snake robots have been developed. 

Camarillo et al. developed a forward and inverse kinematics mapping for a tendon-driven cardiac catheter based on constant curvature assumption \cite{camarillo2008mechanics}. Rucker and Webster provided a static and dynamic model for a continuum robot with general tendon routing using Cosserat-rod and Cosserat-string models \cite{rucker2011statics}. Yasin and Simaan developed a position/force controller for a continuum robot using support vector regression to cancel fiction losses, followed by a least-squares model optimization to minimize residual joint-force errors \cite{yasin2021joint}. Most of these model-based analytical approaches are very effortful for modeling highly nonlinear kinematics of continuum robots. Modeling and identification of such mechanical properties as friction or hysteresis could be further challenging using these approaches. 

To avoid these limitations, several model-less approaches are developed for online learning of robot kinematics \cite{yip2014model, fagogenis2016adaptive, grassmann2018learning, thamo2021hybrid}. But, most of these model-less approaches are based on online numerical estimation of robot kinematics and impose a high computational load of solving nonlinear numerical optimizations and differential equations, which is restrictive for real-time control applications. 

Moreover, even a higher level of positioning accuracy ( $\sim25 \mu m$) is required to perform a safe operation close to the retina \cite{vander2020robotic}. Since there are always mechanical assembly issues, especially related to the wire's initial tension that causes nonlinear properties such as hysteresis in the snake robot's behavior, it would be difficult and costly to reduce hysteresis in the hardware, but as an alternative, hysteresis can be modeled and efficiently compensated for by the software. Therefore, having a model that compensates for the effect of hysteresis is highly required and can significantly improve robot control accuracy.  

In this paper, we present a probabilistic data-driven kinematic model with a hysteresis compensation algorithm. We employ a probabilistic model, Gaussian Mixture Model (GMM) – Gaussian Mixture Regression (GMR) \cite{sylvain2009robot}, for nonlinear mapping between the input signals and the outputs (snake bending angles, yaw, and pitch). The contributions of this paper consist of:
\begin{itemize}
    \item This probabilistic model can learn the input-output relation of the robot kinematics in the GMM phase via optimizing the model parameters using the Expectation Maximization (EM) algorithm and then predict the output for any given input in the GMR phase in a fast, smooth, and robust way.
    \item We proposed a new method in the GMM training phase by splitting each data cycle into two main clusters and applying the GMM algorithms once on the main cycle and once again on the two separate clusters and developed a new algorithm that searches between these three clusters and solves the inverse optimization problem with higher precision compared to a nominal model.        
    
    \item Moreover, we incorporate a hysteresis compensation algorithm into the probabilistic kinematics model which significantly improves the model accuracy as compared to the same model without hysteresis compensation, and also to a prior model developed by \cite{song2017intraocular}. 
    
\end{itemize}

The remainder of this paper is organized as follows. In Section \ref{sec: Material_Method}, we introduce the I$^2$RIS robot mechanism, the experiment setup for data collection, the image processing method for robot configuration detection, and the mathematics of the robot kinematics model. Experimental results are provided in Section \ref{sec: Results}. Conclusion and future direction are discussed in Section \ref{sec: Conclusion}.

\section{Material and Method} \label{sec: Material_Method}

\subsection{System Description}

\begin{figure}[t!]
    \centering
    \includegraphics[width=0.49\textwidth]{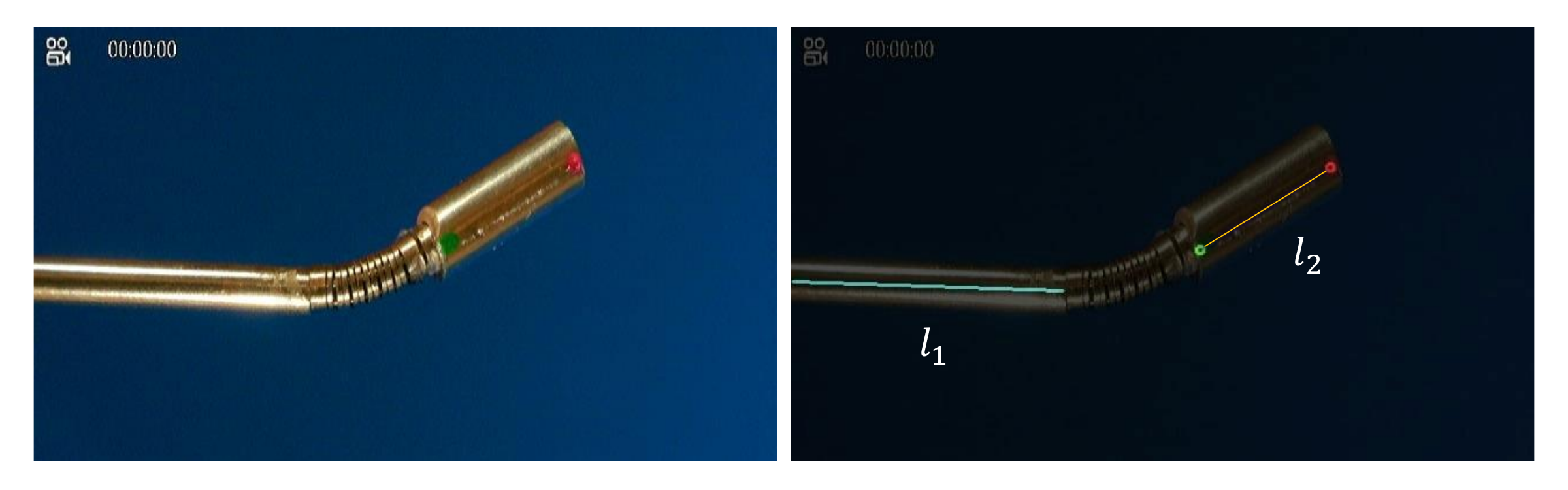}
      \caption{Vision-based bending angle detection. (Left) Original image captured by the camera, (Right) Detected target points, the line connecting them ($l_2$) and robot base orientation line ($l_{1})$. The angle between $l_{1}$ and $l_{2}$ provides the bending angle.}
       \vspace{-0.5cm}
      \label{fig: VisionBasedAngleDetection}
   \end{figure}

We developed a 3-D calibration setup for accurate orientation measurement of the integrated robotic intraocular snake (I$^2$RIS) robot (Fig. \ref{fig: experimet_setup}). The robot has two degrees of freedom (DOFs) capable of providing $\pm45\degree$ for pitch and yaw motions. It is equipped with a miniature gripper at the tip of the dexterous distal section with $\phi0.9$ mm in diameter and 3.0 mm in length actuated by four wires $\phi$0.15 mm made of SUS 304 \cite{jinno2022microgripper}. The actuation unit comprises two DC servo motors (DCX08M EB KL 4.2V, Maxon Motor Inc.) with reduction gears (GPX08 A, gear ratio 64:1), two encoders (ENX 8 MAG 256IMP counts/turn), and controller (EPOS2 24/2). The instrument unit is easily detachable from the actuation unit which makes it desirable for sterilization (Fig. \ref{fig: I2RIS_model}) \cite{jinno2020improved}. 

 The main software for the low-level control of the Maxon motor control library for the snake actuation unit is developed in C++, but the high-level motion control algorithm and the vision-based configuration detection are developed in Python and the communication between them is performed through ROS. 
 


\subsection{Vision-Based Measurement}

To detect the robot's bending angle, a set of image processing techniques is employed (Fig. \ref{fig: experimet_setup}) using the OpenCV library of Python \cite{howse2016opencv}. For this purpose, a small cylinder ($\phi0.9 \times \phi1.4 \times 4.2 mm$), colored with green and red dots at its two ends, is mounted on the robot tip as shown in (Fig. \ref{fig: VisionBasedAngleDetection}). To be more robust to lighting conditions for color filtering, the video frames are converted from RGB color space to HSV space. Afterward, appropriate color thresholds are employed to mask the green and red colors,  as two ends of the cylinder. Followed by a median operator on the masked pixels, two ends of the cylinder are identified. The bending angle, the relative angle between the line connecting the cylinder ends $l_2$ and the base shaft $l_1$, is then estimated as the angle between $l_1$ and $l_2$ (Fig. \ref{fig: VisionBasedAngleDetection}). 

Two digital microscopes (AD407 Andonstar, China) are connected to the computer via USB interface to capture the real-time top and side view videos with a magnification ratio up to $270\times$ and resolution of $4032 \times 3024$ with a scale factor equal to 20 $\mu m/pixel$. The microscopes are calibrated in advance using the Camera Calibrator app of MATLAB. 



\subsection{Mathematical Formulation} \label{sec: Math}

\subsubsection{Forward kinematics}

\begin{figure}[t!]
    \centering
    \includegraphics[width=0.45\textwidth]{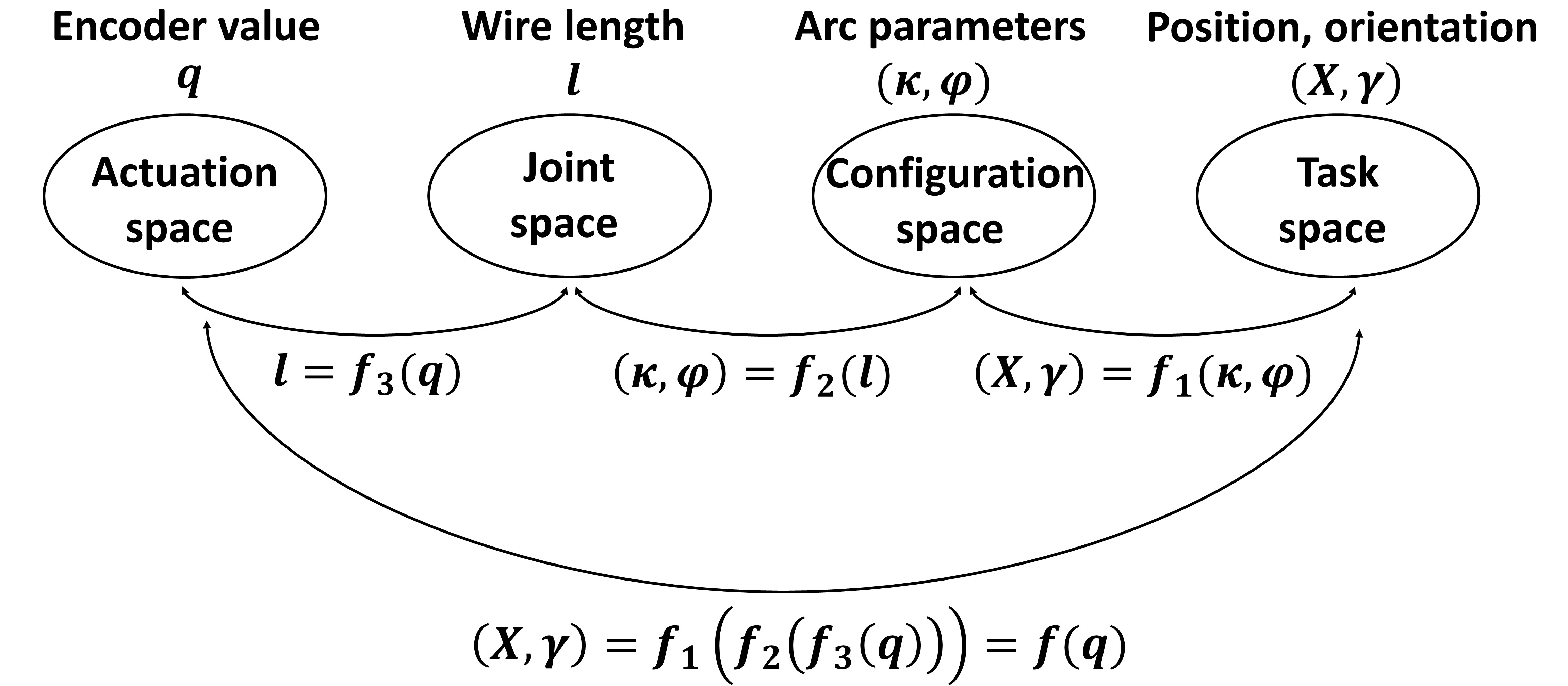}
      \caption{ Forward kinematic schematic: a mapping between the actuation space, the joint space, the configuration space, and the task space. }
      \label{fig: kinematic_mapping}
   \end{figure}

Forward kinematics modeling is important for the control of snake robots. It could be performed by doing a mapping between the actuation space (encoders), the joint space (wire length), the configuration space (arc parameters), and the task space (snake position and orientation) (Fig. \ref{fig: kinematic_mapping}). We developed an experimental forward kinematic mapping between the actuation space and the task space, for we cannot measure the joint and configuration space variables of the robot.  
   
The total dataset collected during nine reciprocating movement cycles over the snake range of motion is defined as follows,
\begin{equation}
\mathcal{D} =  \left[\begin{matrix}
q_{1,1} & q_{1,2} & \cdots & q_{c,n} &  \cdots & q_{C,N}   \\ 
\gamma_{1,1}  & \gamma_{1,2} & \cdots & \gamma_{c,n} & \cdots & \gamma_{C,N} 
\end{matrix}\right] \in \mathbb{R}^{D \times \left( C\times N \right)} 
 \label{eq: data_set}
\end{equation}
in which $\gamma_{c,n}$ is the snake bending angle for $n^{th}$ data point of the $c^{th}$ test cycle, $q_{c,n}$ is the control input for the $n^{th}$ data point of the $c^{th}$ cycle, $C$ is the total number of cycles, $N$ is the number of data points in each cycle, $\mathbb{N} = C\times N$ is the total number of data points for all training cycles, $D=2$ is the dimension of data points, and $j^{th}$ data point $(j=1,2,\ldots,\mathbb{N})$ is defined as follows, 

\begin{equation}\label{eq: data_point_sai_j}
\xi_j  =
\begin{bmatrix}\xi_j^I\\ 
\xi_j^O
\end{bmatrix} =
\begin{bmatrix}q_j\\
\gamma_j
\end{bmatrix}
\in\mathbb{R}^{{D}\times{1}}
\end{equation}
where super scripts $I$ and $O$ denote input and output, respectively. Using the Gaussian Mixture Model (GMM), the probability for a datapoint $\xi_j$ belonging to the GMM including $K$ Gaussian distributions is defined as follows, 
\begin{equation}
\begin{split}
{\mathcal{P}(\xi}_j) &=\sum_{k=1}^{K}{\mathcal{P}\left(k\right)\mathcal{P}(\xi_j|k)} \\ 
\mathcal{P}\left(k\right)&=\pi_k \\ 
\mathcal{P}\left(\xi_j\middle| k\right)&=\mathcal{N}\left(\xi_j\middle|\mu_k,\mathrm{\Sigma}_k\right)\\
&=\frac{1}{\sqrt{{(2\pi)}^D|\mathrm{\Sigma}_k|}}e^{-\frac{1}{2}\left({(\xi_j-\mu_k)}^T{\mathrm{\Sigma}_k}^{-1}\left(\xi_j-\mu_k\right)\right)}
\end{split}
\label{eq: Gaussian_definition_GMM}
\end{equation}
in which $\mathcal{P}\left(k\right)=\pi_k$ is a prior probability, $\mathcal{P}(\xi_j|k)$ is the conditional probability density, $\mu_k$ and $\mathrm{\Sigma}_k$ are the mean value and covariance matrices of the $k^{th}$ Gaussian distribution $\mathcal{N}\left(\xi_j\middle|\mu_k,\mathrm{\Sigma}_k\right)$. 
Then, we used Expectation Maximization (EM) algorithm over the dataset $\mathbf{D}$ to iteratively optimize the parameters of the GMM $(\Theta_k=\{{\pi_k\in\mathbb{R},{\ \mu}_k\in\mathbb{R}^2,\ {\ \mathrm{\Sigma}}_k\in\mathbb{R}^{2\times2}}\}_{k=1}^K)$ subject to the following constraint,
\begin{equation}
\sum_{k=1}^{K}\pi_k=1 \quad  , \quad     \pi_k\in\left[0,1\right]  
\label{eq: GMM_pi_constraint}
\end{equation}
\vspace{-0.5pt}
After training the GMM model with EM algorithm \cite{moon1996expectation, alexander_fabisch_2021_4889867} and optimizing the parameters $\Theta_k$, it is now possible to estimate the output (snake bending angle) for any given control input using Gaussian Mixture Regression (GMR). The conditional probability of output $\xi^O$ for a given input $\xi^I$ could then be estimated in the GMR phase as,
\begin{equation}
\mathcal{P}\left(\xi^O\middle|\xi^I\right)\sim\sum_{k=1}^{K}{h_k\mathcal{N}({\hat{\xi}}_k,{\hat{\mathrm{\Sigma}}}_k)}
\label{eq: GMR_equation}
\end{equation}
in which $\hat{\xi}_k$ and $\hat{\mathrm{\Sigma}}_k$ are the expected mean values and covariance matrices of the $k^{th}$ Gaussian distribution and are calculated as follows, 
\begin{equation}
\begin{split}
    \hat{\xi}_k &= \mu_k^O + \Sigma_k^{OI} (\Sigma_k^{OI})^{-1} (\xi^I-\mu_k^I) \\
\hat{\Sigma}_k &=\Sigma_k^O-\Sigma_k^{OI}(\Sigma_k^{I})^{-1} \Sigma_k^{IO}
\end{split}
\label{eq: saiHatK_sigmaHatK}
\end{equation}
and $h_k=\mathcal{P}(k|\xi^I)$ specifies the probability of the $k^{th}$ Gaussian distribution being responsible for $\xi^I$ and is calculated as 
\begin{equation}
h_k=\frac{\mathcal{P}(k)\mathcal{P}(\xi^I|k)}{\sum_{k=1}^{K}{\mathcal{P}(i)\mathcal{P}(\xi^I|i)}}=\frac{\pi_k\mathcal{N}(\xi^I;\mu_k^I,\mathrm{\Sigma}_k^I)}{\sum_{k=1}^{K}{\pi_i\mathcal{N}(\xi^I;\mu_i^I,\mathrm{\Sigma}_i^I)}}
\label{eq: h_k}
\end{equation}
\vspace{-1pt}
Finally, it is possible to approximate the conditional expectation of the output $\xi^O$ for a given input $\xi^I$ using a single Gaussian distribution function $\mathcal{N}(\hat{\xi},\hat{\mathrm{\Sigma}})$ such that \cite{sylvain2009robot} 
\begin{equation}
\begin{split}
    \hat{\xi}=\sum_{k=1}^{K}{h_k{\hat{\xi}}_k} \quad , \quad 
\hat{\mathrm{\Sigma}}=\sum_{k=1}^{K}{h_k^2{\hat{\mathrm{\Sigma}}}_k}
\end{split}
\label{eq: saiHat_sigmaHat}
\end{equation}

\subsubsection{Hysteresis compensation}
\begin{figure}[t!]
    \centering
    \includegraphics[width=0.43\textwidth]{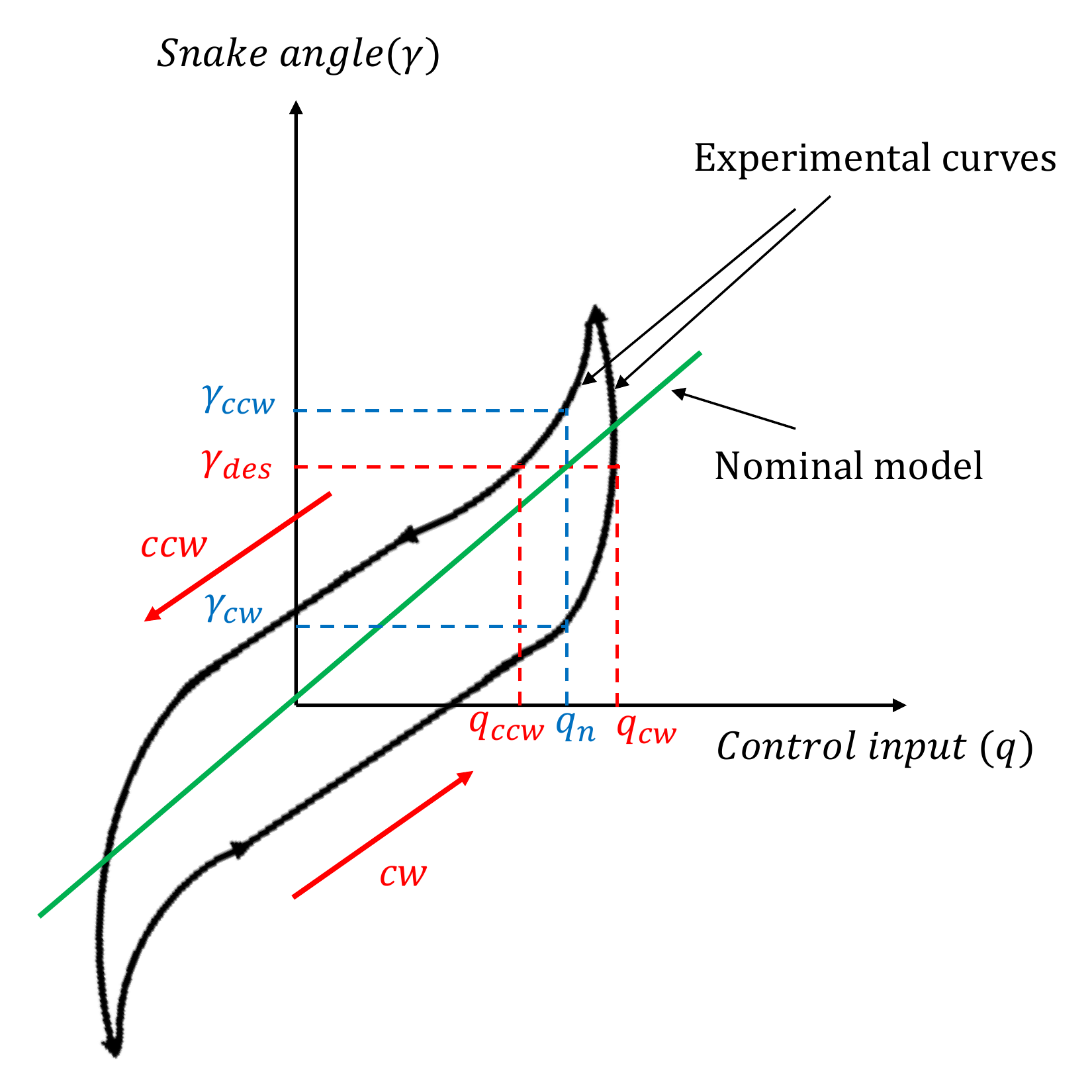}
      \caption{Schematic of the hysteresis loop. Solid black curves represent the proposed model for the actual behavior of the robot depending on the robot actuator direction, clockwise (cw) or counterclockwise (ccw), and the green curve shows the nominal model. }
      \label{fig: Hys_loop}
   \end{figure}
\vspace{-1pt}
Due to the existence of a hysteresis loop in the robot motion, caused by wires tension condition, there could be two possible control inputs $q_{cw}$ and $q_{ccw}$ for any desired snake angle $\gamma_{des}$ (pitch or yaw), and two possible snake angles $\gamma_{cw}$ and $\gamma_{ccw}$ for any nominal control input $q_n$ (Fig. \ref{fig: Hys_loop}). We proposed a method to consider the hysteresis behavior in robot kinematics. Solid black curves (Fig. \ref{fig: Hys_loop}) represent a schematic of the actual behavior of the robot depending on the robot actuator direction, clockwise (cw) for the ascending curve and counterclockwise (ccw) for the descending curve, whereas the green curve shows nominal linear or nonlinear models without considering the hysteresis loop, such as the one developed by \cite{murphy2014predicting, azimi2017teleoperative}. To do the hysteresis compensation we divided the modeling procedure into three main phases. In the first phase, we provided the input-output data collected in the entire reciprocating motion cycle of the snake robot employing the GMM/GMR algorithm (\ref{eq: data_set}-\ref{eq: saiHat_sigmaHat}) to generate the nominal model (the green line of Fig. \ref{fig: Hys_loop}). In the second phase, we split the experimental data of each reciprocating cycle of the robot motion into two clusters, clockwise (cw) for the ascending curve and counterclockwise (ccw) for the descending curve, and repeated the GMM/GMR algorithm (\ref{eq: data_set}-\ref{eq: saiHat_sigmaHat}) for these labeled data (the directed black curves of Fig. \ref{fig: Hys_loop}). From now on, the subscripts cw and ccw are used for the trained model over the ascending and descending data sets, respectively (note that the notations cw and ccw have to do with the motor shaft direction rather than the hysteresis loop direction itself which in our data collection experiment happened to be counterclockwise given the provided input sequence to the robot). Therefore, the parameters of the GMM algorithm for the aforementioned three curves are optimized using the EM algorithms and represented as $\Theta_j^k=\{{\pi_j^k,{\ \mu}_j^k,\mathrm{\Sigma}}_j^k\}_{k=1}^K$ where the subscript $j\in\{n, cw, ccw\}$ stands for the nominal, ascending (cw) and descending (ccw) curves, respectively. Finally, the optimum solution for the robot inverse kinematics model with hysteresis compensation is found by solving the following constrained optimization problem through the iterative search algorithm \ref{alg: hys_loop_model} (see below) which makes use of the nominal, cw and ccw models as follows, 
\begin{equation}
\begin{aligned}
q^*  &= \operatorname*{argmin}_q \frac{1}{2}||\gamma_{des} - \gamma_{GMR}||^2 \\  
  \textrm{s.t.} \quad &  q_{min}\leq q \leq q_{max}   \quad \quad \quad \\
\end{aligned}
\label{eq: inverse_optimization}
\end{equation}
\begin{algorithm}
\caption{Inverse Kinematics with Hysteresis Compensation }\label{alg: hys_loop_model}
\begin{algorithmic}
\State Initialise: $q_{prev} \gets 0$ 
\Require $q_{min}\leq q^{i} \leq q_{max}$
\Ensure for any $\gamma_{des}$, compute the corresponding input  $q_n=\hat{\xi}(1)$ using iteration on the nominal model   \eqref{eq: saiHatK_sigmaHatK}-\eqref{eq: saiHat_sigmaHat} 
 
\State $i \gets 0$
\State $q^i \gets q_{prev}$
\State $\gamma_{GMR}^i \gets \hat{\xi}(q^i)$ \quad  \eqref{eq: saiHat_sigmaHat} 

\While{$||\gamma_{des} - \gamma_{GMR}^{i}||\geq \epsilon$}
\If{$(\gamma_{GMR}^i < \gamma_{des}) \& (q_i > q_{prev})$}
    \State $q^{i+1} = q^i + \alpha \times\big(\gamma_{des}-\hat{\xi}_{cw}^i(2)\big)$
    \State $\gamma_{GMR}^{i+1} \gets \hat{\xi}_{cw}^{i+1}(2)$
    \State $i \gets i+1$ 
    \Comment{compute the corresponding input \State \quad \quad \quad \quad \quad \quad \quad \quad   $q_{cw}=\hat{\xi}_{cw}(1)$ by iteration on  \State \quad \quad  \quad \quad \quad \quad \quad \quad  the clockwise curve} 
\ElsIf{$(\gamma_{GMR}^i < \gamma_{des}) \& (q_i < q_{prev})$}
    \State $q^{i+1} = q^i + \alpha \times\big(\gamma_{des}-\hat{\xi}_{ccw}^i(2)\big)$
    \State $\gamma_{GMR}^{i+1} \gets \hat{\xi}_{ccw}^{i+1}(2)$
    \State $i \gets i+1$ 
    \Comment{compute the corresponding input \State \quad \quad \quad \quad \quad \quad \quad \quad   $q_{ccw}=\hat{\xi}_{ccw}(1)$ by iteration on\\ \quad \quad \quad \quad \quad \quad \quad \quad \quad \quad \quad \quad  \quad  the counterclockwise curve} 
    \ElsIf{$(\gamma_{GMR}^i > \gamma_{des}) \& (q_i > q_{prev})$}
    \State $q^{i+1} = q^i + \alpha \times\big(\gamma_{des}-\hat{\xi}_{cw}^i(2)\big)$
    \State $\gamma_{GMR}^{i+1} \gets \hat{\xi}_{cw}^{i+1}(2)$
    \State $i \gets i+1$ 
    \Comment{compute the corresponding input \State \quad \quad \quad \quad \quad \quad \quad \quad   $q_{cw}=\hat{\xi}_{cw}(1)$ by iteration on \State \quad   \quad \quad \quad \quad \quad \quad \quad   the clockwise curve} 
\ElsIf{$(\gamma_{GMR}^i > \gamma_{des}) \& (q_i < q_{prev})$}
    \State $q^{i+1} = q^i + \alpha \times\big(\gamma_{des}-\hat{\xi}_{ccw}^i(2)\big)$
    \State $\gamma_{GMR}^{i+1} \gets \hat{\xi}_{ccw}^{i+1}(2)$
    \State $i \gets i+1$ 
    \Comment{compute the corresponding input \State \quad \quad \quad \quad \quad \quad \quad \quad   $q_{ccw}=\hat{\xi}_{ccw}(1)$ by iteration on \State \quad   \quad \quad \quad \quad \quad \quad \quad   the counterclockwise curve} 
\EndIf
\State $q_{prev}  \gets q^{i+1}$ 
\EndWhile
\end{algorithmic}
\end{algorithm}
\vspace{-1pt}
where the step-size $\alpha$ is updated using the Armijo rule. 

\section{Results} \label{sec: Results}

\begin{figure}[t!]
    \centering
    \includegraphics[width=0.51\textwidth]{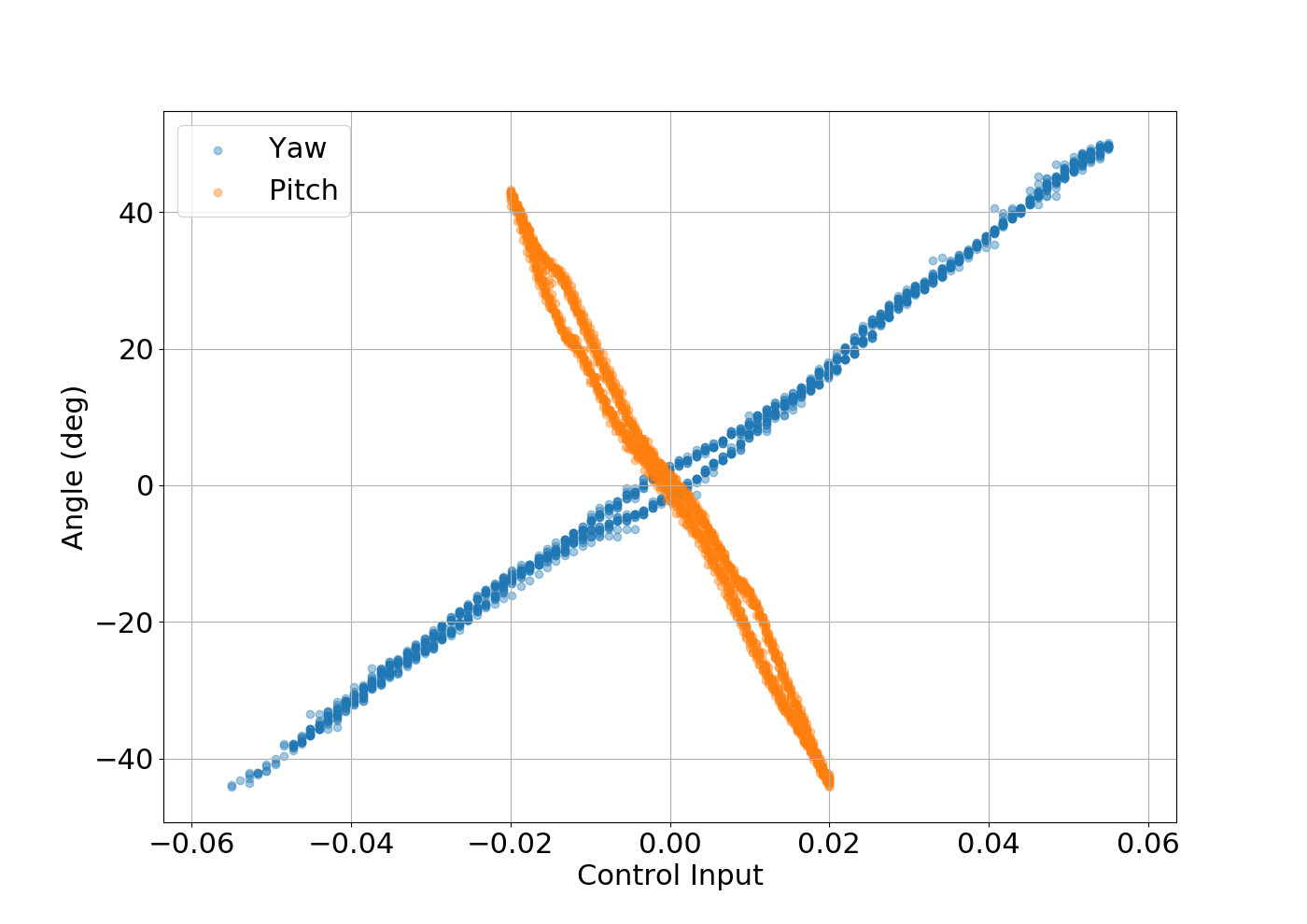}
      \caption{Experimental result of the I$^2$RIS robot, control input signal vs. bending angle (deg) measured by microscope for top view (Yaw) and side view (Pitch).}
      \label{fig: Yaw_Pitch_raw}
   \end{figure}

\begin{figure}[t!]
    \centering
   \subfigure[]{
    \centering
    \includegraphics[width=0.51\textwidth]{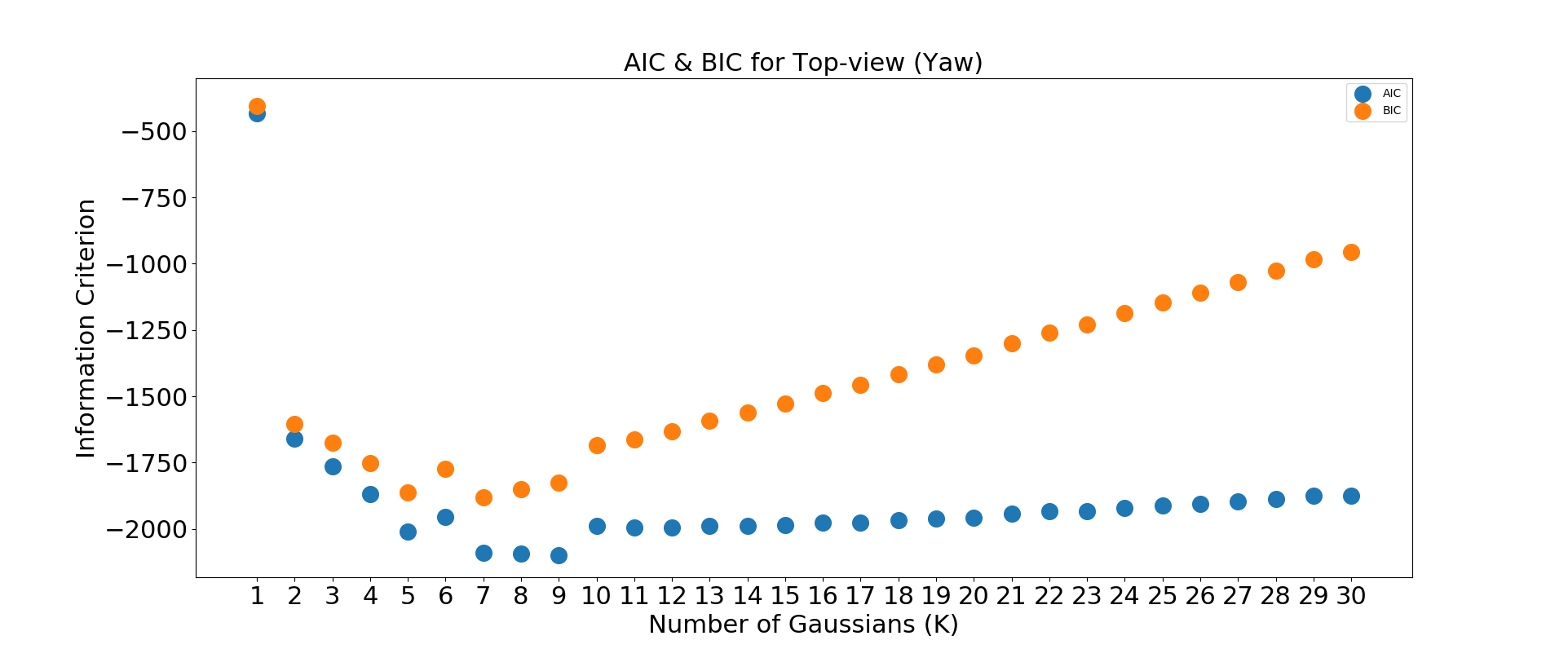}
    \label{fig:BIC_Yaw}}
    \subfigure[]{
    \centering
    \includegraphics[width=0.51\textwidth]{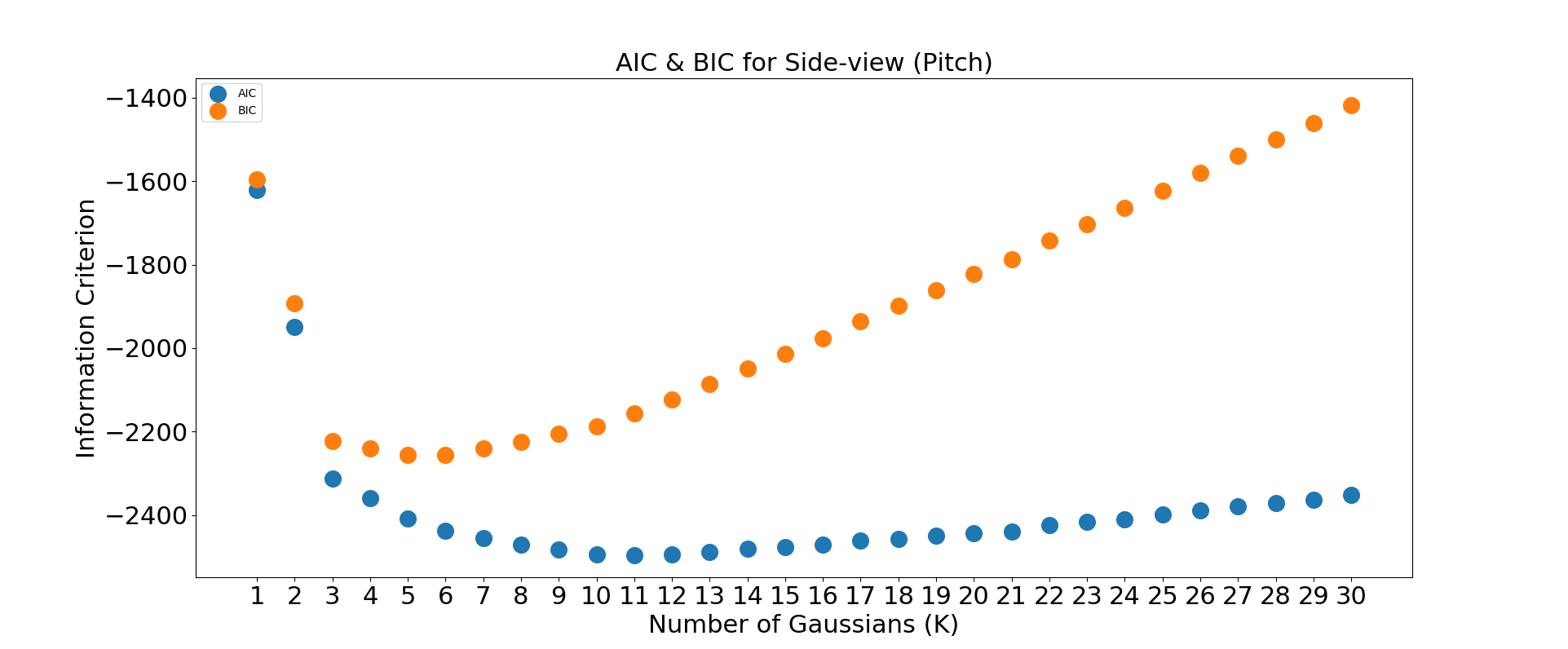}
    \label{fig:BIC_Pitch}}
      \caption{BIC and AIC values vs. number of Gaussian distributions. }
      \label{fig: BIC_AIC}
   \end{figure}

To collect the experimental data set, we activated the yaw motion of the snake robot from its initial home position (straight configuration) by having the robot move in discrete and defined steps (200 steps for a complete reciprocating cycle), pausing for 3 sec after each step and capturing an image with the top-view microscope. Then, we repeated the same scenario for the pitch motion and collected images in the same way. For each motion, we collected nine reciprocating cycles, of which six cycles were used during the training phase and three cycles were used for the prediction or test phase. Experimental results show that the pitch motion has a wider hysteresis loop than the yaw motion (Fig. \ref{fig: Yaw_Pitch_raw}). Tackling this nonlinear behavior necessitates considering hysteresis compensation for improving the robot model accuracy and control performance. This results in improvements in both, the safety and precision, of robot-assisted microsurgery. 
In the first phase, representing nominal model training, in order to select the optimum number of Gaussian distributions, we used Bayesian Information Criterion (BIC) and Akaike Information Criterion (AIC) to estimate the optimum value of $K$ which is a tradeoff between the prediction accuracy and the model complexity (Fig. \ref{fig: BIC_AIC}). For example, according to the AIC measure, the initial estimate of the optimum number of $K$ for yaw and pitch is 9 and 11, respectively. The results of the nominal model error for three different tests are listed in Table \ref{tab: result_table}. The best accuracy reached by only using the nominal model during the test phase was $1.73\degree$ for the pitch and $1.01\degree$ for yaw.   

\begin{figure*}[t!]
	\centering
	\subfigure[]{
    \centering
    \includegraphics[width=0.48\textwidth]{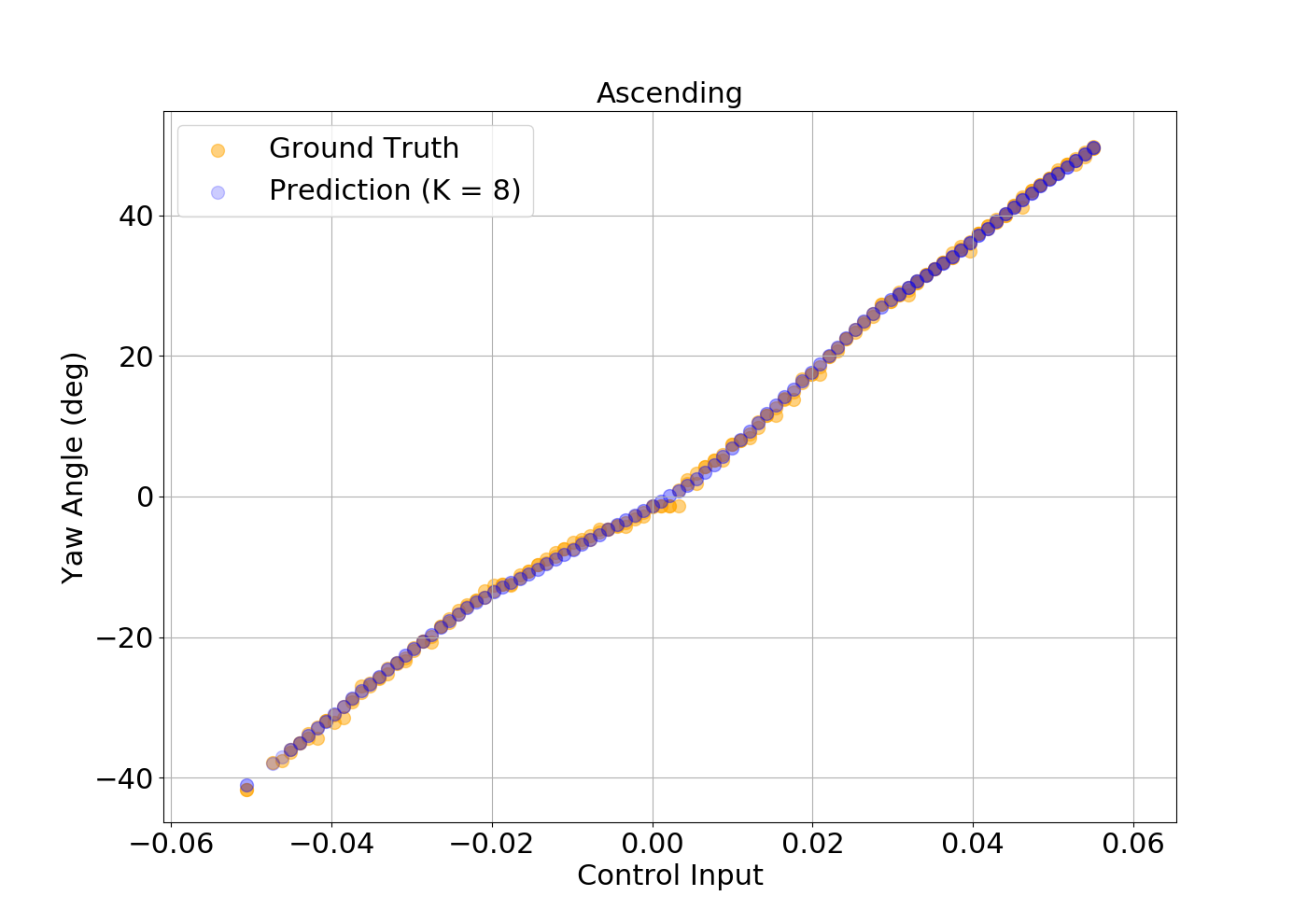}
    \label{fig:Top_view_Yaw_ascending}}
    \subfigure[]{
    \centering
    \includegraphics[width=0.48\textwidth]{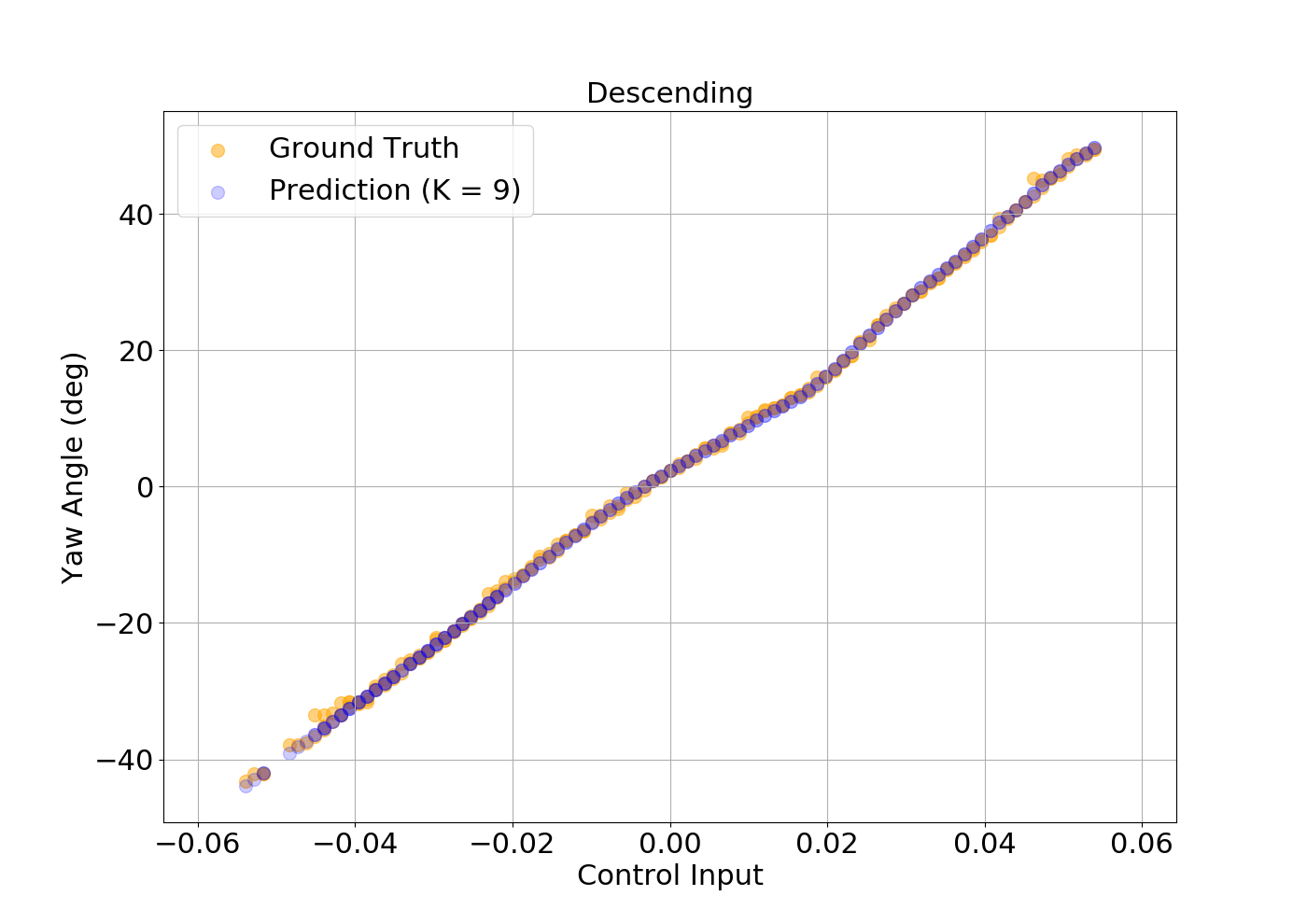}
    \label{fig:Top_view_Yaw_descending}}
    \subfigure[]{
    \centering
    \includegraphics[width=0.48\textwidth]{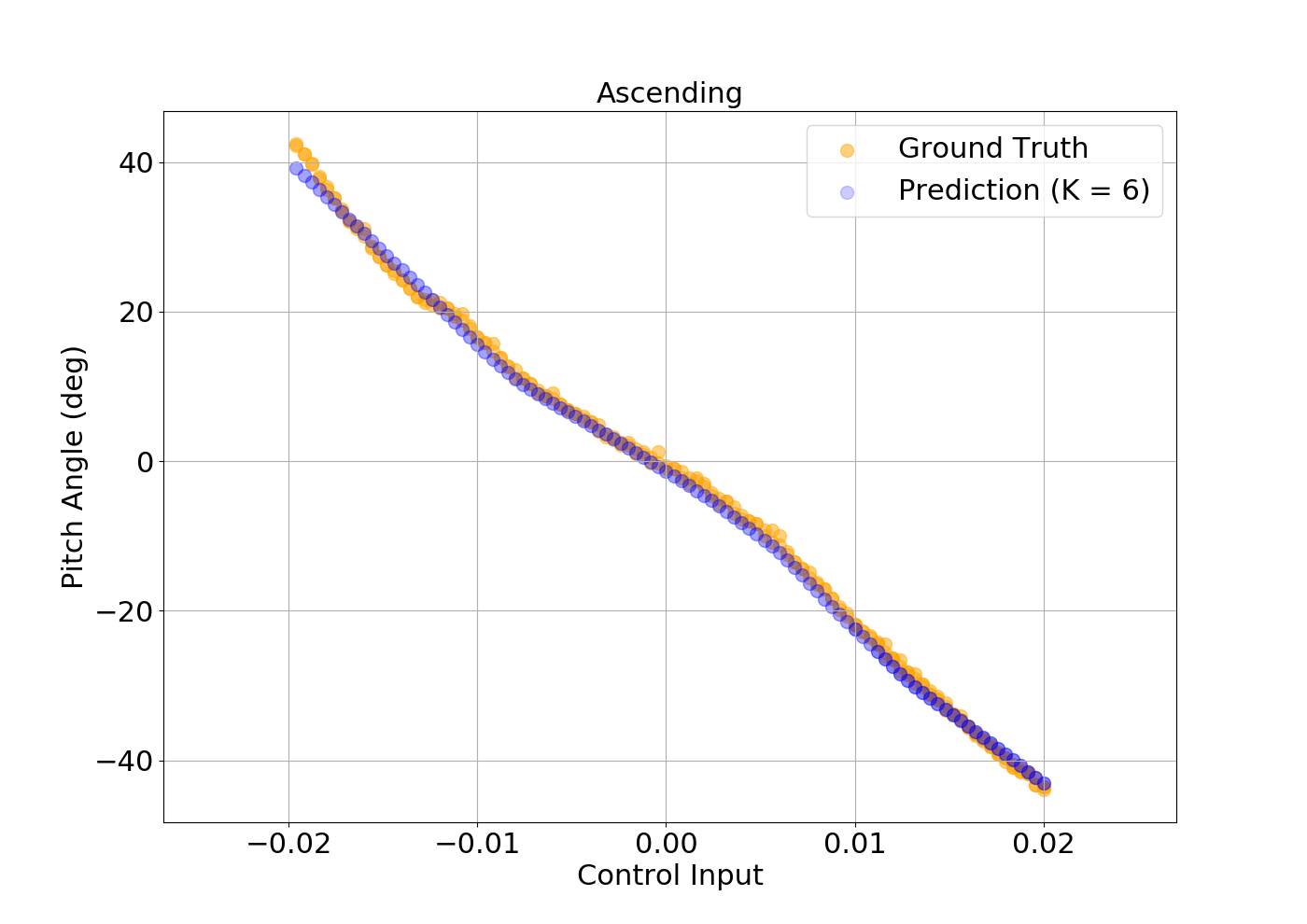}
    \label{fig:Side_view_Pitch_ascending}}
    \subfigure[]{
    \centering
    \includegraphics[width=0.48\textwidth]{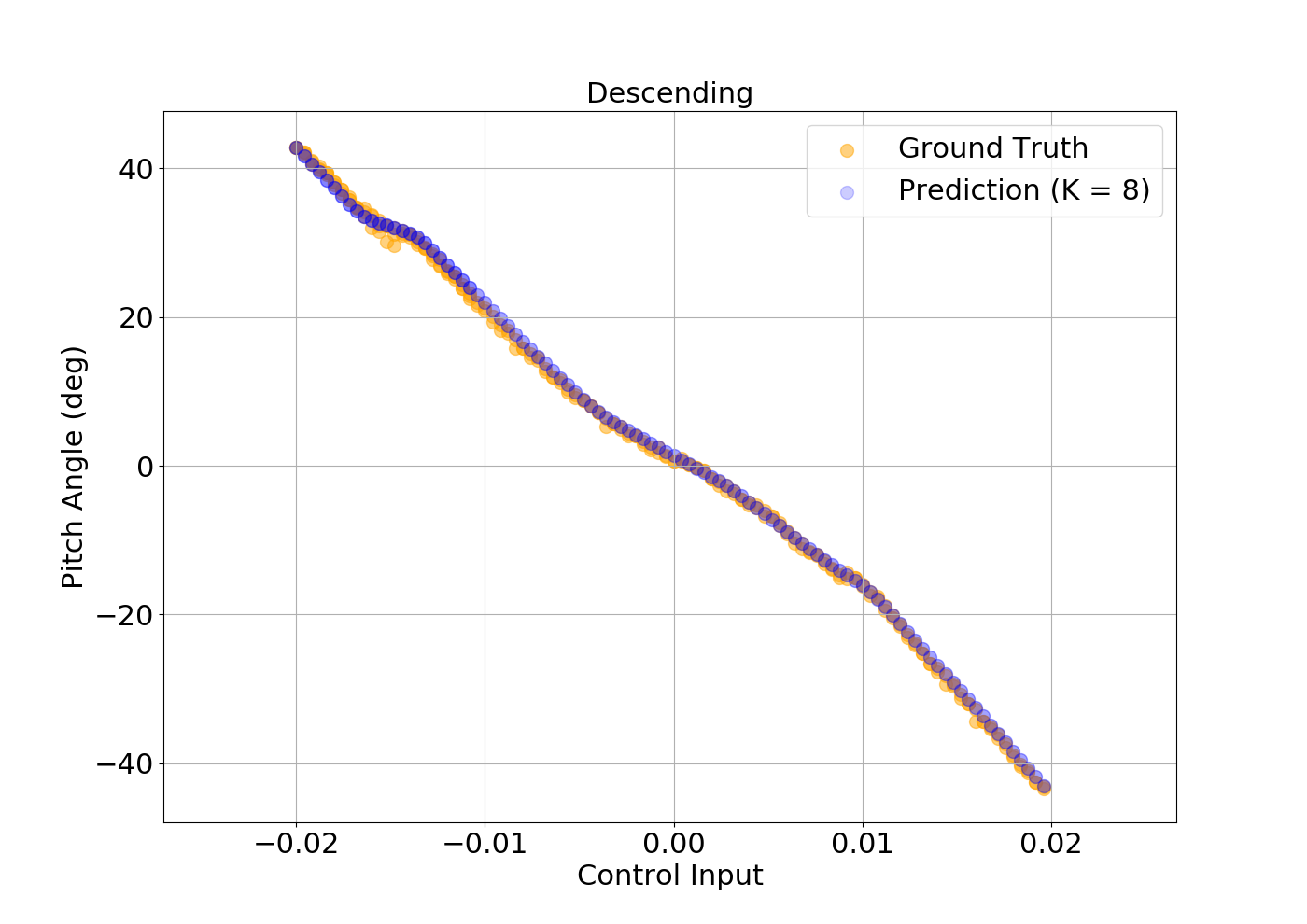}
    \label{fig:Side_view_Pitch_descending}}
    \caption{Second phase of the model training by separating the ascending (cw) and descending (ccw) curves for yaw and pitch motions; (a) yaw ascending curve, (b) yaw descending curve, (c) pitch ascending curve, (d) pitch descending curve.}
     \vspace{-0.5cm}
\label{fig: TopView_SideView_raw_2}
\end{figure*}

In the second phase of model training, we split the motion cycle of each yaw and pitch angle into two clusters, i.e., the ascending (cw) and descending (ccw) curves, and trained two separate models for each curve over six cycles (12 half-cycle curves for each of yaw and pitch motions) and trained the model based on this data set (Fig. \ref{fig: TopView_SideView_raw_2}).       

\begin{figure*}[t!]
	\centering
	\subfigure[]{
    \centering
    \includegraphics[width=0.48\textwidth]{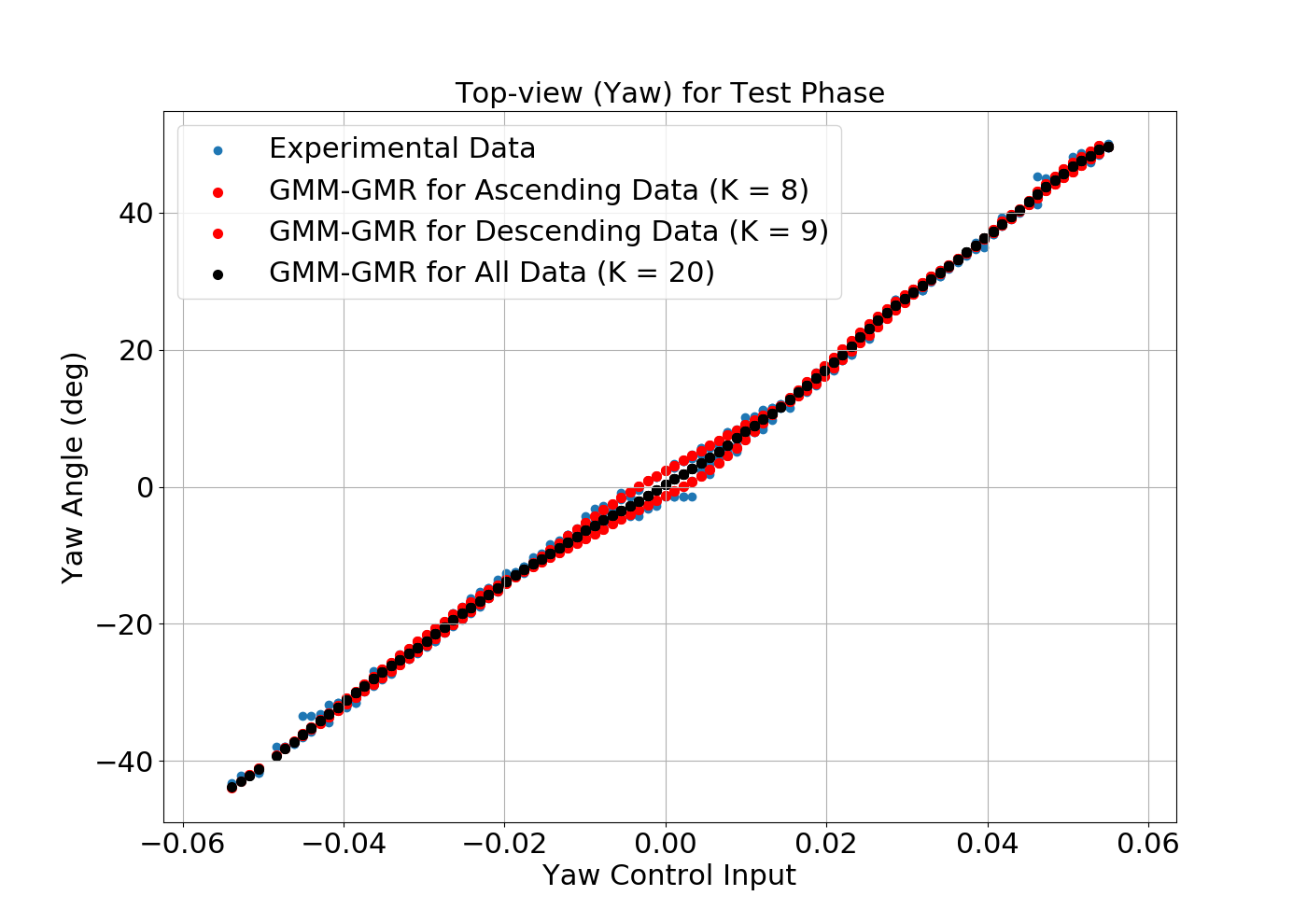}
    \label{fig:Top_view_Yaw_GMR}}
    \subfigure[]{
    \centering
    \includegraphics[width=0.48\textwidth]{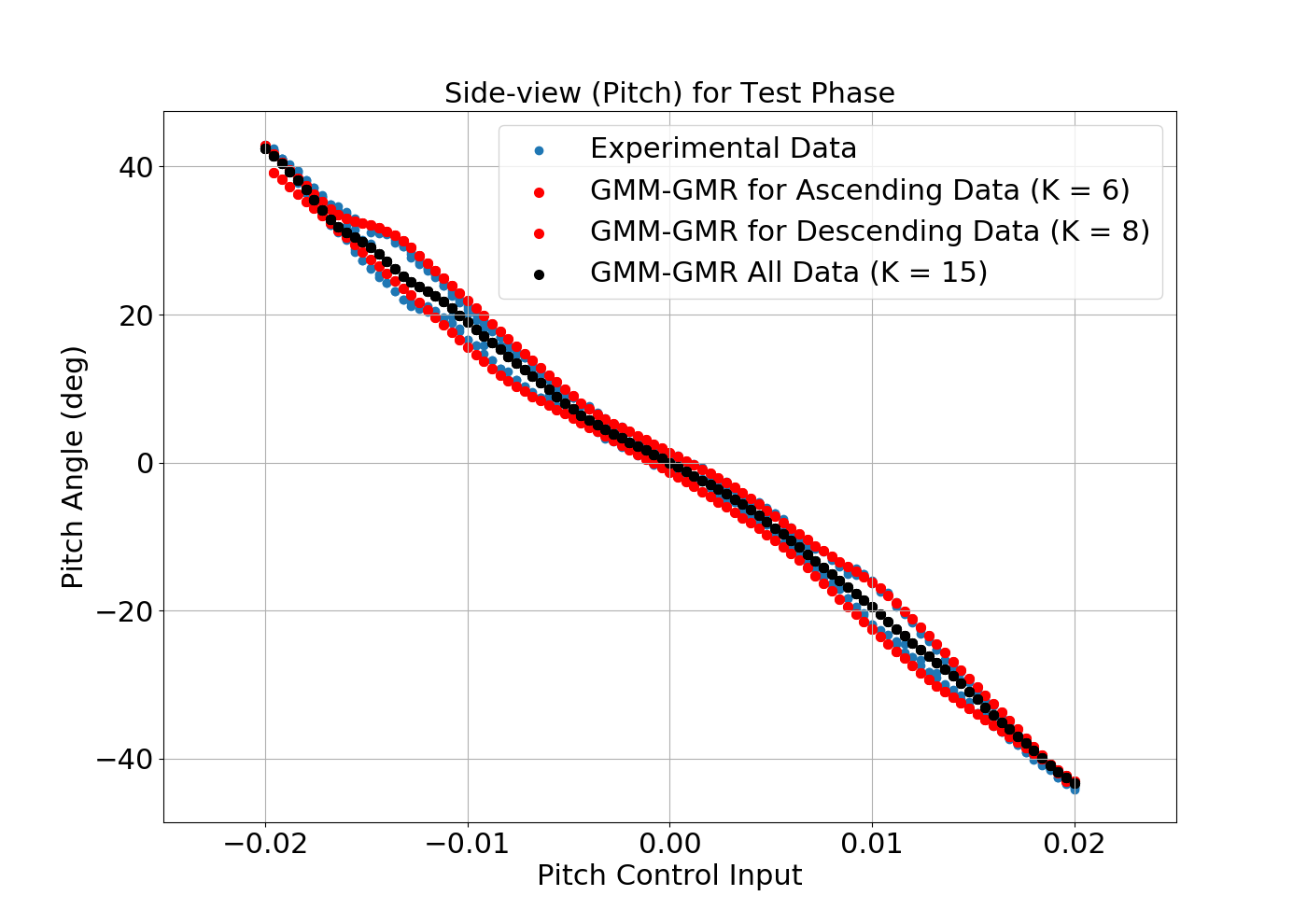}
    \label{fig:Side_view_Pitch_GMR}}
    \caption{Results of the nominal model (solid black curve) and the proposed model with hysteresis compensation algorithm (red dots) during the test phase over three remaining motion cycles are shown for, (a): top view (yaw), (b): side view (pitch).}
     \vspace{-0.5cm}
\label{fig: TopView_SideView_GMR}
\end{figure*}

Finally, by implementing the hysteresis compensation model using algorithm \ref{alg: hys_loop_model} on the trained nominal, ascending, and descending models, we were able to improve the model accuracy relative to the nominal model by a significant amount. Fig. \ref{fig: TopView_SideView_GMR} illustrates the result of the proposed model during the test phase on the rest of the three cycles in which the nominal model (black curve) and the two ascending (cw) and descending (ccw) models are used to predict the experimental output (red curves). The results demonstrate that the proposed model with hysteresis compensation considerably reduced the angle RMSE of yaw and pitch motion by 61$\%$ and 73$\%$, respectively, as compared to the nominal model without hysteresis compensation (\ref{tab: result_table}). Given the 3 mm length of the snake, these errors are equivalent to 92 $\&$ 53$\mu m$ tip position errors for pitch and yaw motions if using the nominal model, which is not safe, whereas, the proposed model has 24 $\&$ 20$\mu m$ errors for pitch and yaw which is an acceptable positioning accuracy for retinal procedures \cite{vander2020robotic}.


 \vspace{-1pt}

\begin{table}[t!]
\centering
    \caption{RMSE (deg) of the nominal and proposed models for 3 different tests.}
    \includegraphics[width= 0.44\textwidth]{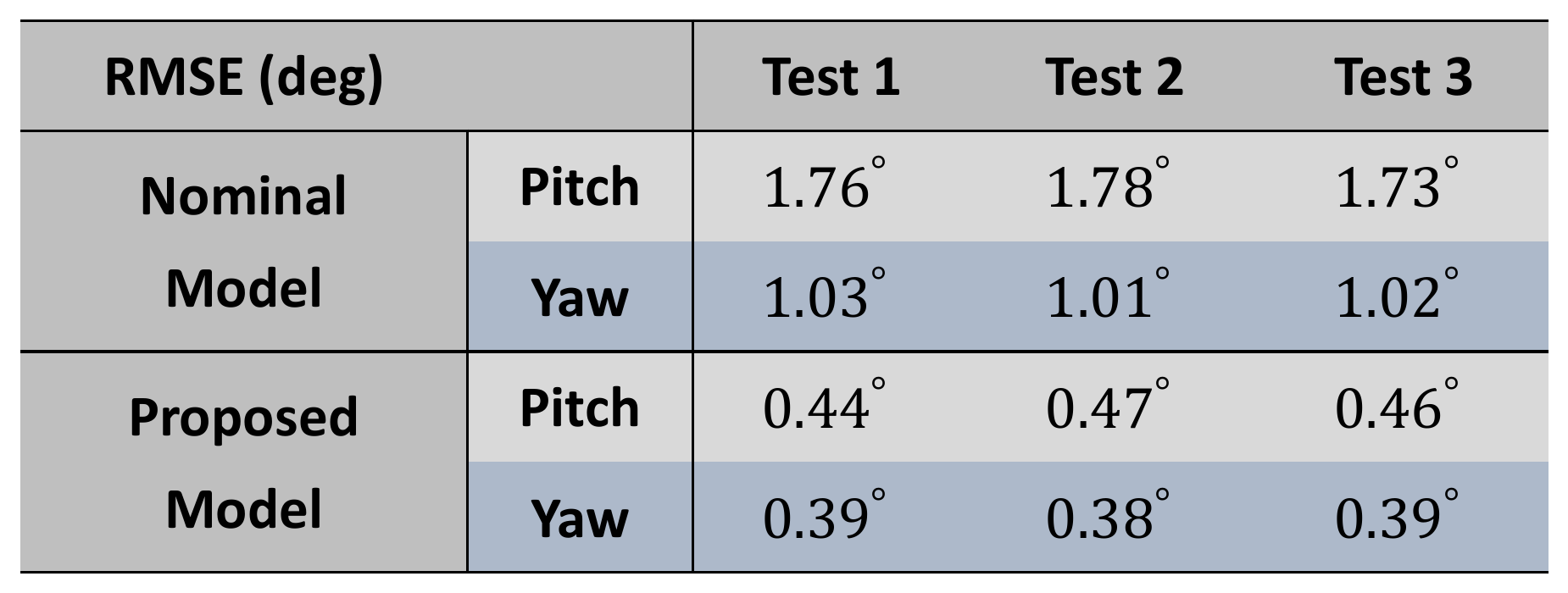}
      \label{tab: result_table}
       \vspace{-0.5cm}
\end{table}

\section{CONCLUSIONS} \label{sec: Conclusion}
In this work, we employed a data-driven probabilistic GMM-GMR model that demonstrated to be capable of learning the nonlinear features of the I$^2$RIS snake robot kinematics during the GMM training phase and reproducing the robot output during the GMR phase for the test dataset with 1.73$\degree$ and 1.01$\degree$ accuracy for pitch and yaw motions, respectively. It is shown that this accuracy was significantly improved by incorporating this model with the proposed hysteresis compensation algorithm to reach $\sim$0.4$\degree$ of accuracy, which, based on our clinical lead, is acceptable.

The accuracy improvement for the pitch motion is greater than that for the yaw motion ($\sim12\%$) due to its wider hysteresis loop (Fig. \ref{fig: Yaw_Pitch_raw}), therefore, adding the hysteresis compensation algorithm to the I$^2$RIS robot is justifiable and is recommended to improve its accuracy and to compensate for the accuracy reduction caused by wire tension drops resulting from overuse of the snake robot.   

Future work includes an evaluation of the effectiveness of the proposed model in a closed-loop kinematic control framework of the robot. It is considered that the current data-driven forward kinematic model is developed by robot motion in free space without external loads. Therefore, its precision may be predicted to drop significantly if the robot has interreacted with tissue which may increase the load on the tool. In order to compensate, we intend to incorporate an FBG sensor into the snake robot for shape and force sensing and to develop an adaptive control framework to improve the robot's control accuracy in the presence of external loads on the robot's body.






\section*{ACKNOWLEDGMENT}

We appreciate Ali Ebrahimi, Trent Tang, and Ji Woong Kim for their technical support to prepare the experiment setup. We are also grateful to Prof. Makoto Jinno for all of his efforts to design and fabricate the robot.


\bibliographystyle{IEEEtran}
\bibliography{root}

\end{document}